\documentclass[11pt,a4paper]{article}
\usepackage[hyperref]{emnlp2018}
\usepackage{times}
\usepackage{latexsym}

\usepackage{url}

\usepackage{amsmath,amssymb}
\usepackage{tabularx}
\usepackage{multirow}
\usepackage{arydshln}
\usepackage{color}
\usepackage{caption}
\usepackage{subcaption}
\usepackage{url}
\usepackage{enumitem}
\usepackage{graphicx}
\usepackage{pgfplots}
\usepackage{amsmath}
\usepackage{array}
\newcommand{\newvec}[1]{\mathbf{#1}}

\newcommand{\specialcell}[2][c]{\begin{tabular}[#1]{@{}c@{}}#2\end{tabular}}

\aclfinalcopy 

\title{Zero-shot Transfer Learning for Semantic Parsing}

\author{Javid Dadashkarimi \\
   Computer Science Department\\
  Yale University\\
  {\tt javid.dadashkarimi@yale.edu} \\\And
    Alexander Fabbri \\
  Computer Science Department \\
  Yale University \\
  {\tt alexander.fabbri@yale.edu} \\  \AND
  Sekhar Tatikonda \\
  Department of Statistics and Data Science  \\
  Yale University \\
  {\tt sekhar.tatikonda@yale.edu } \\ \And 
    Dragomir R. Radev \\
  Computer Science Department  \\
  Yale University \\
  {\tt dragomir.radev@yale.edu }
  \\}

\date{}
\begin{document}
\maketitle
\begin{abstract}
  While neural networks have shown impressive performance on large datasets, applying these models to tasks where little data is available remains a challenging problem.
  In this paper we propose to use feature transfer in a zero-shot experimental setting on the task of semantic parsing. 
  We first introduce  a new method for learning the shared space between multiple domains based on the prediction of the domain label for each example. 
  Our experiments support the superiority of this method in a zero-shot experimental setting in terms of accuracy metrics compared to state-of-the-art techniques.
  In the second part of this paper we study the impact of individual domains and examples on semantic parsing performance.
  We use influence functions to this aim and investigate the sensitivity of domain-label classification loss on each example. 
  Our findings reveal that cross-domain adversarial attacks identify useful examples for training even from the domains the least similar to the target domain.
Augmenting our training data with these influential examples further boosts our accuracy at both the token and the sequence level.

\end{abstract}

\section{Introduction}
Transfer learning, or applying features and models trained on one task to another task, is a growing frontier for deep neural models \cite{Ng:2016}.
In a wide range of tasks, the number of training examples is limited, yet we want neural models to be able to match the performance of networks from other tasks in generalizing to unseen data \cite{Vinyals:15,Xian:17,Ravi:16}.
Even though adapting knowledge hidden in one domain is expected to be useful for another domain, investigating the role of each example is computationally expensive \cite{Koh:2017}.

In this paper we study two sides of transfer learning: 1) feature transfer and 2) zero-shot learning.
Our case study is semantic parsing, which can be viewed as a type of multi-class classification\footnote{Zero-shot learning has been primarily studied for single class prediction \cite{Romera:2015,Vinyals:15,Ravi:16}.}. 
Our contribution for the first part is to use a single encoder and a single decoder.
We then minimize our loss function which controls the distance of our model from the gold domain label. 
In other words, we propose to use example-level classification of domain labels. Thus the crucial improvement of this model is its ability to map samples even in different domains into one shared space while utilizing a fewer number of parameters. 
Our experimental results support the superiority of this model compared to state-of-the-art shared-encoder and shared-decoder models in zero-shot settings.
Experiments on two semantic parsing data sets augmented by sequence to sequence training data reveals that in zero-shot settings (i.e., when there are few examples in one class \cite{Xian:17}) adding more parameters for training degrades the accuracy.
We propose to use one matrix, of size the number of tasks by the size of the current hidden dimension, to map each example to its corresponding domain.

For the second part, we apply a robust statistic, the influence function, for each sample and then assess its effect on binary domain classification error. We then study the impact of each domain (removing the domain and re-training) as well as each example (adding noise or adversarial attacks \cite{Koh:2017}) on semantic parsing. 
To this aim we conduct two sets of experiments; 
first we remove and re-train each task and then test the model on one particular task.
We define the closest domain as the domain in which we see the largest drop in accuracy during testing after removing it. We select the closest and farthest domains and use the influence functions for cross-domain adversarial attacks. 
Our results confirm that even tasks with varying topics have useful information to share for semantic parsing.
Finally, our findings also confirm that cross-domain adversarial attacks do not necessarily have uniform effects; adversarial examples from one domain might be more intensive than in the reverse direction, as identified by influential functions.

\section{Previous Work}
\subsection{Semantic Parsing}
Semantic parsing aims to map natural language to formal meaning representations. Previous approaches have used induced grammars to parse the source text to a given logical form \cite{Zettlemoyer:05, Zettlemoyer:07,  Zettlemoyer:2012, Kwiatkowksi:10, Kwiatkowski:11, Kwiatkowski:13}. More recently, neural network models and variants of the sequence to sequence models used in machine translation have been successfully applied to semantic parsing \cite{Dong:2016, Jia:2016, Zhong:2017, Xu:17}. While the majority of semantic parsing has focused on in-domain data, the sparseness of datasets has led to recent work in out-of-domain semantic parsing through domain adaptation, transfer learning and multi-task learning. \citet{Su:17} convert target logical forms from domains in the OVERNIGHT dataset \cite{Wang:2015} to a canonical representation and then learn a paraphrase model from the source to target representation. 


Other work aims to exploit data from multiple knowledge bases and tasks for semantic parsing by varying and combining parsing architectures. \citet{Herzig:17} motivate the use of multiple knowledge bases by observing that the structure of language composition repeats across domains. They propose variants of one-to-one, many-to-many and one-to-many architectures with varying levels of sharing between the encoder and decoder elements. Best results are achieved when parameter sharing across the model is maximum. Additionally, the gain in accuracy for such models is most visible when there is a small amount of data. \citet{Fan:17} observe similar results, although they perform multi-task experiments and observe the effect of training with syntactic parsing for semantic parsing. They hypothesize that the encoder-decoder framework learns a canonical representation across tasks. 
\subsection{Shared Encoder and Decoder Transfer Learning}
As stated above \citet{Herzig:17} compared different encoding-decoding settings for transfer learning between domains. 
In general, transfer learning requires a shared space for mapping dimensions by the following: 1) a shared encoder (one to many) where the encoder is shared between all domains 
or 2) a shared decoder (many to many) where each domain has its own encoder/decoder with an additional shared domain


\citet{Herzig:17} compared these approaches with the proposed single encoder single decoder augmented by a one-hot domain vector and reported higher performance.
However, often introduce a large number of parameters during training or do not proposing sufficient regularization controls over these parameters.
In this paper we formulate this problem with zero-shot learning where each domain is considered a class seen once/never before, and our aim is to find an accurate mapping between a test example and the training network.
\subsection{Zero-shot learning}
By definition, zero-shot learning is a learning approach for predicting examples of one class which has not been explored before. 
Variants of this problem take the form of one-shot or few-shot learning which are less restrictive than the basic definition \cite{Xian:17}.
Let $X\in R^{d\times m}$ be the set of training examples, where $m$ is the number of examples and $d$ is the dimension of each example.
Each example has one or more class labels $z_j$, but in the zero-shot setting, for new examples we predict $z'_j$ where it is likely to be out-of-domain or sparse. 
\citet{Romera:2015} proposed to use a predictor for each example based on either a multi-class or binary classification task. 
\citet{Vinyals:16} proposed to use $ \hat{y} = \sum_{i=1}^{k} a(x,x_i) y_i$ as an estimator for the new coming instance $x$ which uses an attention kernel $a(.,.)$ on $X\times X$ \cite{Goldberger:04}.

\section{Zero-shot Learning for Semantic Parsing}
In this section we introduce our proposed multi-domain transfer learning method.
The main idea is to use information from other domains through the decoding process. 
Let $T=\{t_1,..,t_K\}$ be the set of $K$ tasks and $S_k=\{(\newvec{x}_l,\newvec{y}_l,t_k)\}_{l=1}^{N_{k}}$ the set of $N_{k}$ examples in $t_k \in T$. We can define the probability of generating token $w$ by:
\begin{align*}
    p(y_j=w|x,y_{1:j-1}) &= \\ \sum\limits_{k=1}^{K}p(y_j=w|x,y_{1:j-1},t_k)&p(t_k|x,y_{1:j-1}),
\end{align*}
where the first part of the summation can be estimated by \citet{Jia:2016}:
\begin{eqnarray*}
    p(y_j=w|x,y_{1:j-1},t_k) &\propto \exp (U_w[s_j,c_j]) \\ 
    p(y_j=\textrm{copy}[i]|x,y_{1:j-1},t_k) &\propto \exp (e_{j,i}) 
\end{eqnarray*}
where $U_w$ is a general decoder matrix over $T$; $s_j\in \mathbb{R}^{k\times 2d}$ and $c_j\in \mathbb{R}^{k\times 2d}$ are the decoder and context states respectively.
The case where we copy token $i$ of the tokens from the source sequence is copy$[i]$ and
$e_{j,i}$ is the attention score of the current state $j$ with respect to the source position $i$.

The second part is:
\begin{eqnarray*}
    p(t_k|x,y_{1:j-1}) \propto \exp (W_{t_k}[s_j,c_j])
    \label{eq:word-task}
\end{eqnarray*}
where $W_T \in \mathbb{R}^{K\times 4d}$ is a mapping between the current state and all the domains. 
Eq. \ref{eq:word-task} gives scores over the $k$ domains.

We predict the task for the current decoder state $j$ as follows:
\begin{eqnarray*}
  \hat{k}= \arg\max_k ~W_{t_k}[s_j,c_j].
\end{eqnarray*}
Then we assign $\hat{k}$ for example $x$ and use the corresponding $p(y_j=w|x,y_{1:j-1},t_{\hat{k}})$ for prediction. 

We also define $\hat{t}_{k'}$ as the task predicted for our encoder (i.e., $\arg\max_{k'} W_{t_{k'}}[b_n,c_n]$ where $n=|x|$ or the size of input).
Finally the loss function is:
\begin{eqnarray*}
  \label{eq:loss}
  \mathcal{L} = &-\log p(y_{1:j}) + \frac{1}{2}||t_k-\hat{t}_k||^2+\frac{1}{2}||t_k-\hat{t}_{k'}||^2
\end{eqnarray*}
where the first term is standard negative log likelihood and the second and the third terms penalize misclassifications of the gold task $t_k$.

By following the above formulation our model tries to estimate a distribution over $k$ tasks/topics for each word (tasks are indicated by $t_{1:k}$ for the input example $x_t$ with initial state $h_{t-1}$).


\subsection{Influence Functions}
\label{Influence Functions}
\citet{Koh:2017} introduced influence functions for determining which training points are most responsible for the predictions of a model.
In this section we first shed light on the impact of this idea in the zero-shot setting.
Let $z_j=(x_j,y_j,t_k)$ be a training sample in task $t_k \in T$ and $\textbf{z}=(\textbf{x},\textbf{y},t_k)$ is the batch of samples in $t_k$.
Let $\mathcal{L}(z_j,\theta)$ be the loss for this particular example according to Eq. \ref{eq:loss} and $\frac{1}{n}\sum_{i=1}^n \mathcal{L}(z_i,\theta)$ the empirical loss \cite{Wasserman:2006}.
\par
We define $$\mathcal{L}(\textbf{z},t_k,\theta)=\sum_{z_i \in \textbf{z}} \mathcal{L}(z_i,t_k,\theta)$$ as the loss over the set of examples in $t_k$, $$\hat{\theta}_{-\textbf{z}} = \arg\min_{\theta}\sum_{z_i\notin \textbf{z}}\frac{1}{n}\mathcal{L}(z_i,\theta)$$ the minimizer after removing task $t_k$ from training, and $$\hat{\theta}_{-z_j} = \arg\min_{\theta}\sum_{z_i\neq z_j}\frac{1}{n}\mathcal{L}(z_i,\theta)$$ the minimizer after removing $z_j$ from training.
The goal is to study the impact of the particular domains and examples on the loss by measure the difference in the parameters found from the above minimizations with the $\hat{\theta}$ when all domains/examples are used. 

The influence function tries to estimate this impact of the domain by up-weighting the examples in that domain through the following equation: $$\hat{\theta}_{\epsilon,\textbf{z}} = \arg\min_{\theta}\sum_{z_i \notin \textbf{z}}\frac{1}{n}\mathcal{L}(z_i,\theta) + \epsilon \mathcal{L}(\textbf{z},\theta,t_k)$$ 
and analogously 
$$\hat{\theta}_{\epsilon,z_j} = \arg\min_{\theta}\sum_{z_i\neq z_j}\frac{1}{n}\mathcal{L}(z_i,\theta) + \epsilon \mathcal{L}(z_j,\theta,t_k)$$ to determine the influence of a particular example $z_j$. This form of up-weighting is useful for measuring the impact of domains and examples as opposed to more expensive leave-one-out calculations. Up-weighting with $\epsilon=-\frac{1}{n}$ and removing $z_j$ are two sides of a same coin \cite{Koh:2017,Wasserman:2006}.

Formally, the loss sensitivity for a domain $\textbf{z}$ and for the example $z_j$ can be written as follows:
$$\mathcal{I}_{\uparrow,\mathcal{L}}(\textbf{z},z_{\textrm{test}})=-\nabla_{\theta} \mathcal{L}(z_{\mathrm{test}},\hat{\theta})^TH_{\hat{\theta}}^{-1} \nabla_{\theta} \mathcal{L}(\textbf{z},\hat{\theta})$$ $$\mathcal{I}_{\uparrow,\mathcal{L}}(z_j,z_{\textrm{test}})=-\nabla_{\theta} \mathcal{L}(z_{\mathrm{test}},\hat{\theta})^TH_{\hat{\theta}}^{-1} \nabla_{\theta} \mathcal{L}(z_j,\hat{\theta})$$ \cite{Cook:1980,Koh:2017}. We use the Hessian Vector Product (HVP) iterative method for efficiently calculating the Hessian matrix in the above equations. See the supplementary material for more detail. 



\section{Experiments}

\begin{figure*}[t]
    \centering
    \begin{subfigure}[t]{0.48\textwidth}
        \centering
        \includegraphics[width=\linewidth]{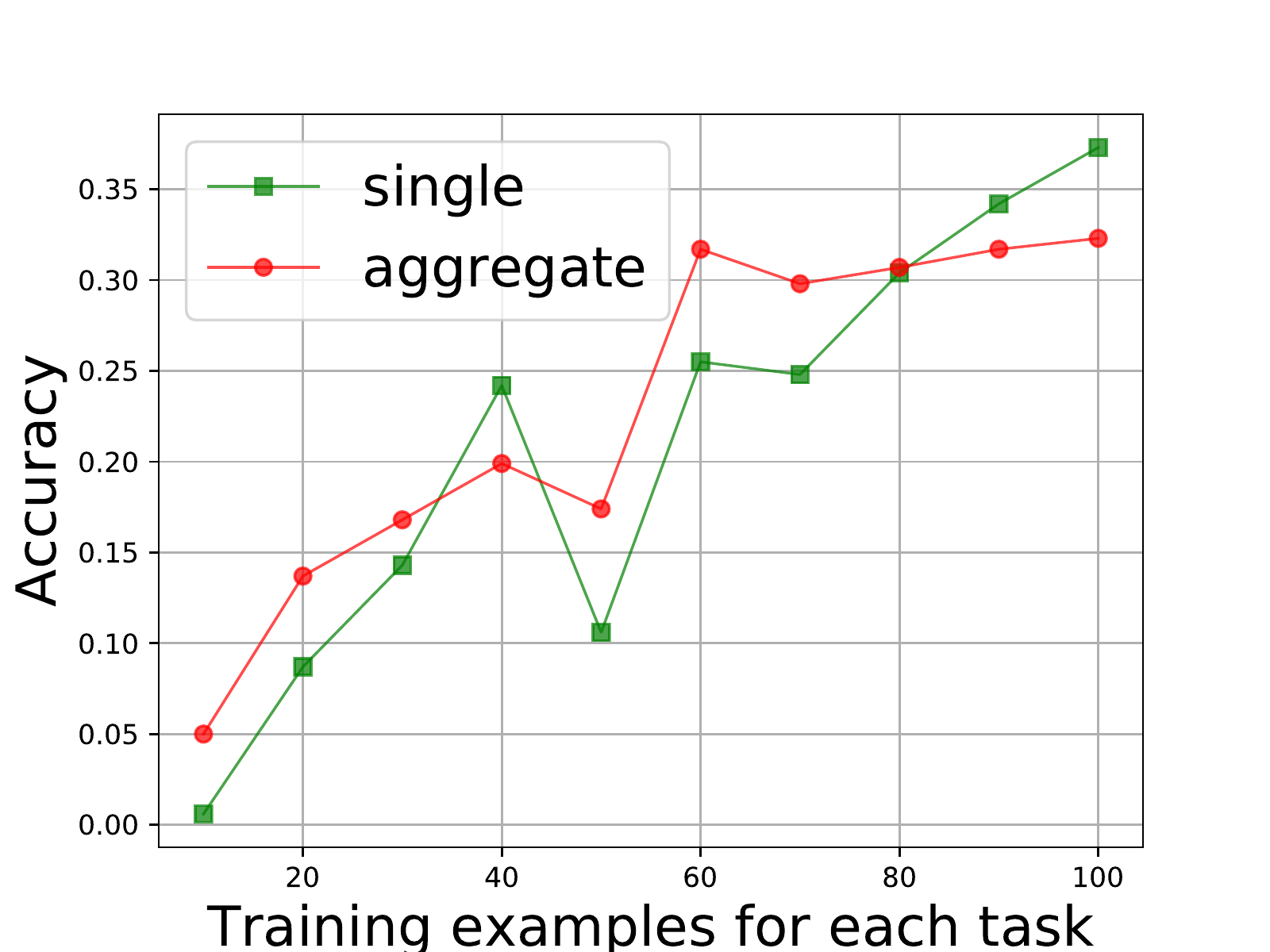}
        \caption{publications}
    \end{subfigure}%
    ~ 
    \begin{subfigure}[t]{0.48\textwidth}
        \centering
        \includegraphics[width=\linewidth]{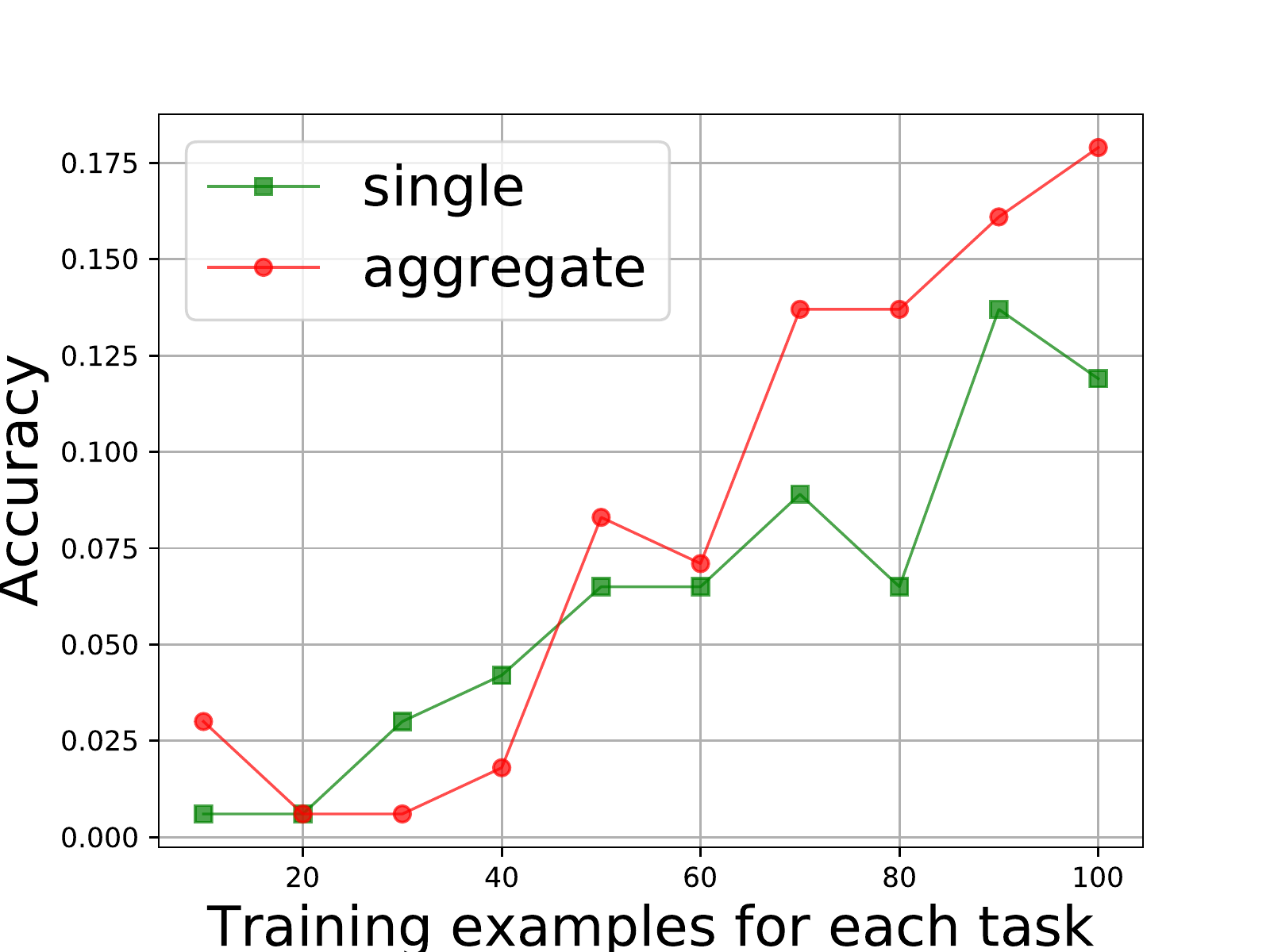}
        \caption{calendar}
    \end{subfigure}
    ~
    \begin{subfigure}[t]{0.48\textwidth}
        \centering
        \includegraphics[width=\linewidth]{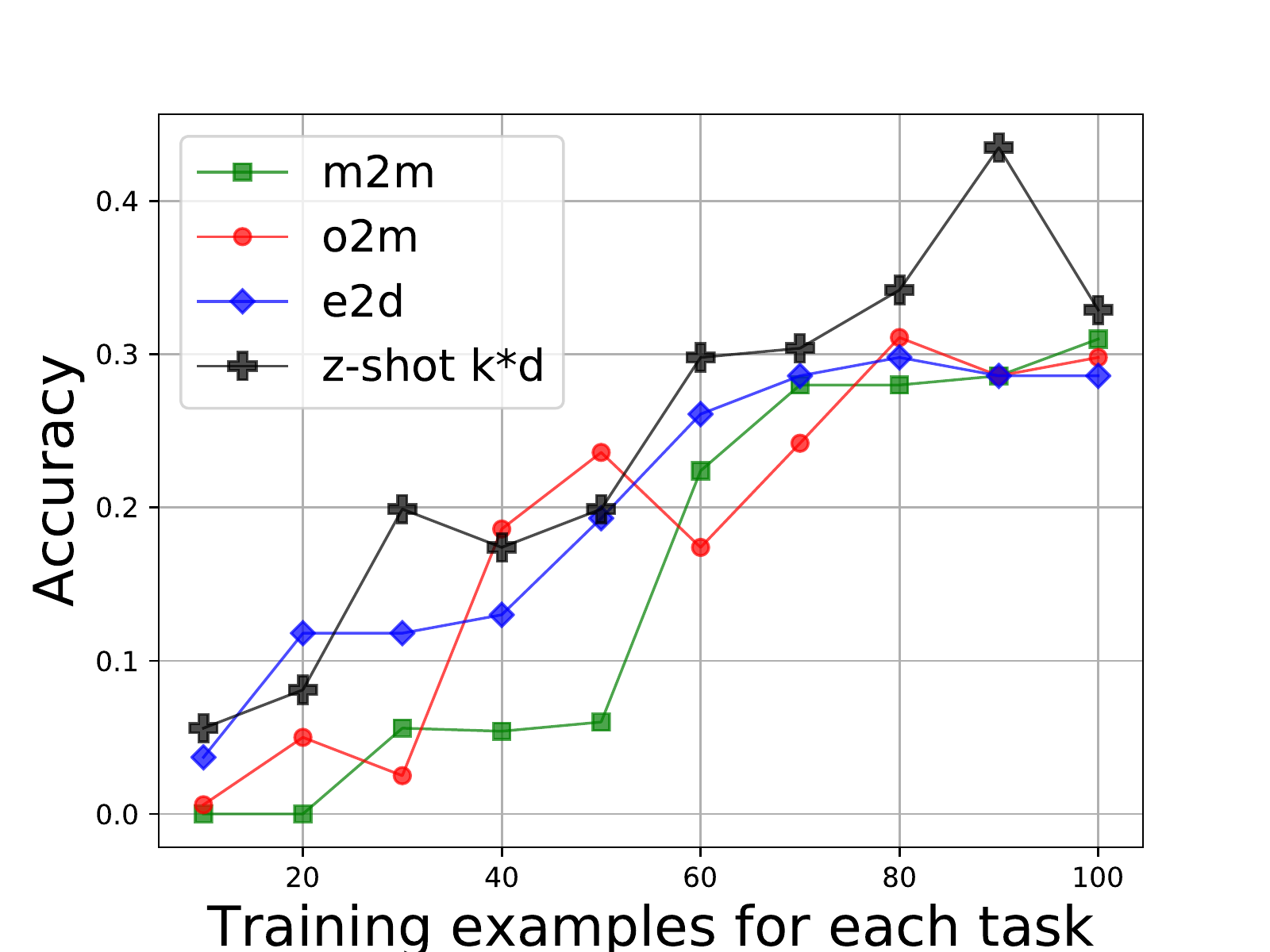}
        \caption{publications}
    \end{subfigure}%
    ~ 
    \begin{subfigure}[t]{0.48\textwidth}
        \centering
        \includegraphics[width=\linewidth]{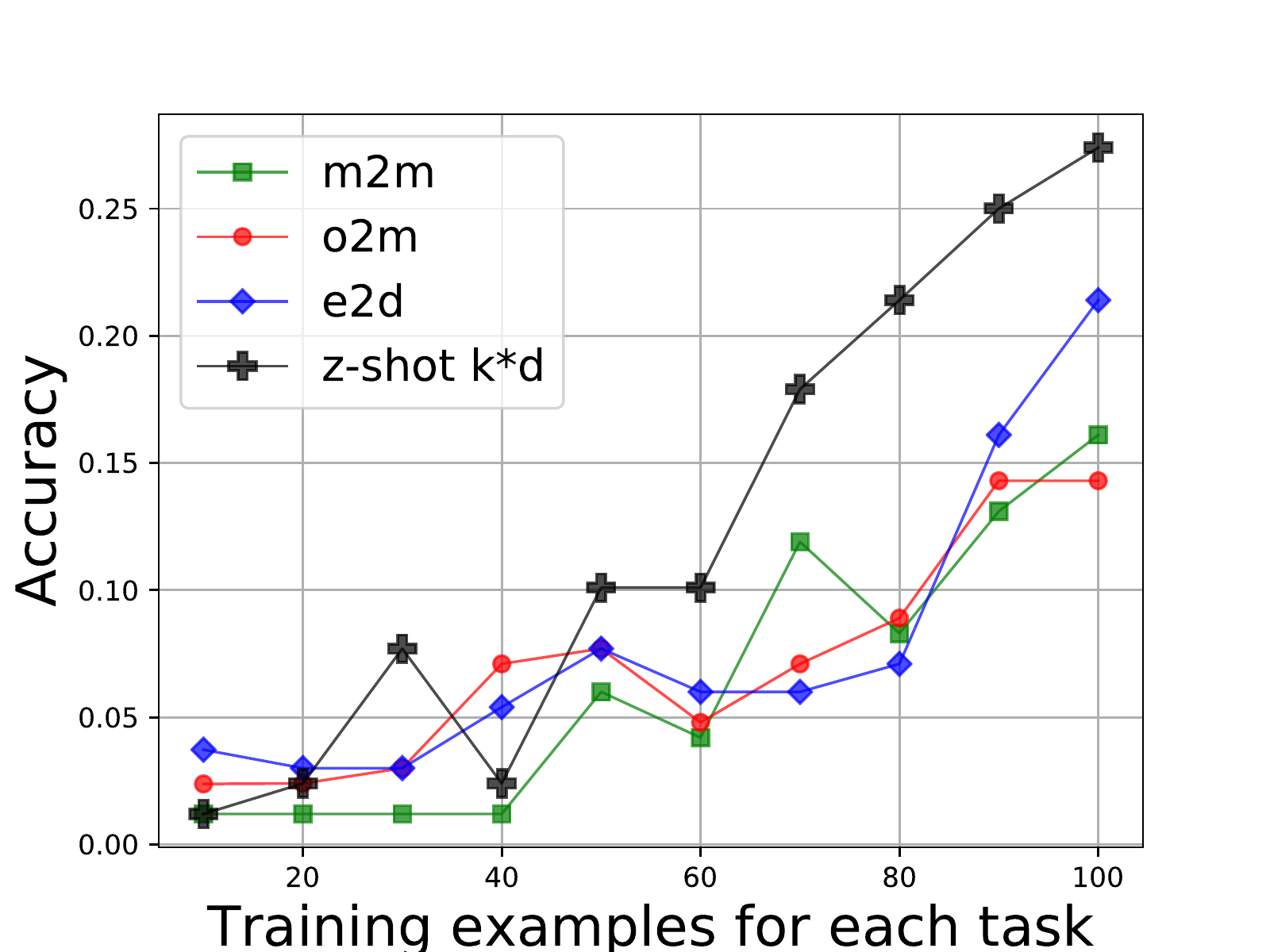}
        \caption{calendar}
    \end{subfigure}
    \caption{Denotation level accuracy for single-task training (i.e., train on task A, test on task A) and multi-task aggregation training (i.e., train on all tasks, test on task A) as a function of training size on the same random shuffle of the data are shown in (a) and (b). A comparison of accuracy for the many to many, one to many, the encoder decoder methods in \cite{Herzig:17} and our proposed zero-shot k*d method as a function of training size are shown in (c) and (d).}
    \label{fig:sing-agg}
\end{figure*}

In this section we report the results of our experiments on multiple multi-task learning approaches for semantic parsing.
We conduct a set of experiments to assess the quality of current approaches for transfer learning when there is limited number of training data in one domain or zero-shot learning settings by definition \cite{Xian:17}.

\subsection{Experimental Settings}
\textbf{Data sets:}
We train our models based on the data sets shown in Table \ref{tab:dataset:seq2logic}. 
We take random small samples in order to study the effect of zero-shot settings. 

\textbf{WIKISQL}:
 WIKISQL consists of a collection of questions and corresponding SQL queries and SQL tables \cite{Zhong:2017}. 

\par
\textbf{GeoQuery}:
GEO  contains natural language queries about U.S. geography and their associated Prolog queries. We follow the prepossessing of \cite{Jia:2016} and use De Bruijn index notation for variable-name standardization \cite{Wong:06}. 
\par
\textbf{ATIS}:
The ATIS dataset contains contains natural language queries to a flights database paired with lambda-calculus queries.

\par
\textbf{Overnight}:
The Overnight dataset contains natural language paraphrases paired with logical forms from eight domains such as restaurants, publications and basketball. The dataset was developed by \citet{Wang:2015} using a crowdsourcing experiment in which Amazon Mechanical Turkers created paraphrases for given logical forms which were generated from a formal grammar.

\textbf{Europarl}
Europarl consists of parallel corpora extracted from the proceedings of the European Parliament and includes versions in 21 languages. It has been used extensively for the task of machine translation. We take $\{10,20,..,100\}$ random samples from the English-French and English-German parallel corpora for our experiments \cite{Koehn:2005}.
\begin{table}[ht]
\centering
\caption{Dataset characteristics.}
\scalebox{0.75}{
\begin{tabular}{|c|c|c|} \hline
ID  & Collection  & \#size \\ \hline
\small{GEOQUERY}  & \specialcell{\citet{Zettlemoyer:2012}} & 880   \\ \hline
\small{ATIS}  & \specialcell{ \citet{Zettlemoyer:2007}} & 5,871\\ \hline
\small{OVERNIGHT}  & \specialcell{\citet{Wang:2015}}& 26,098  \\  \hline
\small{WikiSQL}  & \specialcell{\citet{Zhong:2017}}& 61,297  \\  \hline
French$^{+\textrm{en}}$ & \specialcell{\citet{Koehn:2005}}& 2,190,579	\\ \hline
German$^{+\textrm{en}}$ & \specialcell{\citet{Koehn:2005}}&2,176,537  \\ \hline
\end{tabular}}
\label{tab:dataset:seq2logic}
\end{table}


\textbf{Parameter settings:} We set the hidden size of our models to $|h_t|=200$ and the input embedding dimension to $|x_t|=100$.
We use Long Short-Term Memory (LSTM) units as the basic encoder/decoder unit in all of our experiments.
We initialize all the variables with a uniform distribution in $[-1,1]$ and use stochastic gradient descent as our learning method with a $0.5$ learning rate decreasing by half at each epoch.
We set the number of epochs to $15$ for all the experiments.
In the decoding process we assume a maximum length of $100$ for all the data sets and stop decoding after predicting the end-of-sentence indicator.


\subsection{Results and Discussions}
\begin{table*}[t]
\centering
\begin{tabular}{|c|c|c|c|c|}\cline{1-5}
& method & seq-level & tok-level & den-level \\\hline
\multirow{3}{*}{ calendar } & single & 0.012 & 0.422054 & 0.065 \\ \cline{2-5}
& enc-enc & 0.030 & 0.409097 & 0.077 \\\cline{2-5}
& z-shot & \textbf{0.042} & \textbf{0.572} & \textbf{0.101} \\\cline{1-5}
\multirow{3}{*}{ publication } & single & 0.025 & \textbf{0.574} & 0.106 \\ \cline{2-5}
& enc-enc & 0.068 & 0.448032 & 0.193 \\\cline{2-5}
& z-shot & \textbf{0.087} & 0.57038 & \textbf{0.199} \\ \cline{1-5}
\end{tabular}
\caption{Accuracy of semantic parsing approaches over two domains for models with standard single-domain encoder-decoder, shared encoder-decoder over multiple domains and our proposed z-shot method. For test we used all 168 test examples for \texttt{calendar} and 161 examples for \texttt{publications}.}
\label{tab:res_table:seq2logic:overnight}
\end{table*}

First we conducted a set of experiments on two random domains from the OVERNIGHT collection: \textbf{calendar} and \textbf{publications}. Results are presented in terms of accuracy with respect to the gold standard for the following metrics: sequence-level (the percent of output sequences which fully match the gold standard sequence), token-level accuracy (the percent of output tokens matching the gold tokens) and denotation-level accuracy (the percent of output which when executed against an engine such as a database yields the correct answer). We selected our test domains from the OVERNIGHT collection in particular since we can label the batch of examples $\textbf{z}$ with their domain labels while using similar logical forms, an important point for us since we focus on transferring semantic features rather than syntactic ones. We focus our comparison of models on these datasets in order to assess the generalization ability of our models within the zero-shot setting.

We only consider other collections including Geoquery, ATIS, WikiSQL, and Europarl as additional sources for training.
It is clear that some of these collections are noisy for the test tasks, and thus we also try to assess the quality of the current models in filtering irrelevant data.

During learning we take samples of the same size for each task from $\{10,20,..,100\}$.
Fig. \ref{fig:sing-agg} shows the learning curves for both single-task and multi-task learning methods on these domains.
The single-task (red) lines belong to the experiments with the same domain for training and test examples.
As shown in the figure, both publications and calendar benefit from other tasks (see the aggregate lines) even though some may be considered off-topic examples.
The first intuition is that the source sequences in these collections are English sentences and thus are useful for the encoder. 
The second reason is that these samples are structured as questions (for the Overnight, Geo, and the ATIS collections).
This allows the decoder to find meaningful trends through the training data which is in a similar format to the test examples which follow (\texttt{publications} and \texttt{calendar}).

It is worth mentioning that in some tasks (e.g., publications) the model trains fast and becomes stable after a fewer number of examples (see the single line in publications and calendar).   
On the other hand we see overfitting on the training samples (drops in $50$ and $70$ in publications and calendar).

Table \ref{tab:res_table:seq2logic:overnight} show the results of the proposed method and the augmented shared encoder-decoder by \citet{Herzig:17} with $50$ training examples in each domain.
Both of the methods further boost the performance compared to the single domain training examples. 
Our method achieves more robust results compared to \cite{Herzig:17} and Fig. \ref{fig:sing-agg} compares the results when we use a different number of training data.

\subsubsection{Leave-one-out}
\begin{figure*}[t]
    \centering
    ~
    \begin{subfigure}[t]{0.48\textwidth}
        \centering
        \includegraphics[width=\linewidth]{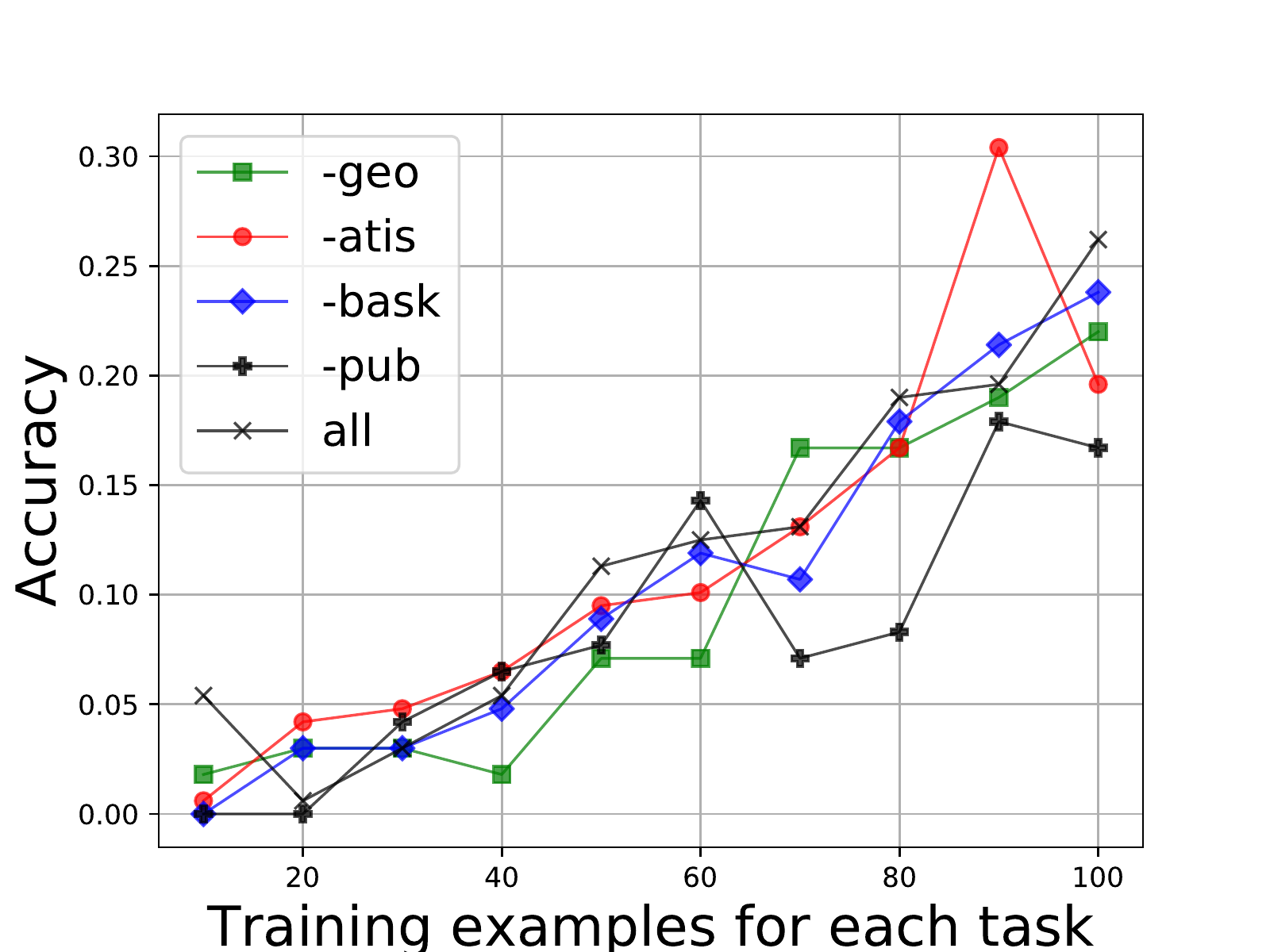}
    
        \caption{den-level}
            \label{fig:lout-den}
    \end{subfigure}
       ~
    \begin{subfigure}[t]{0.48\textwidth}
        \centering
        \includegraphics[width=\linewidth]{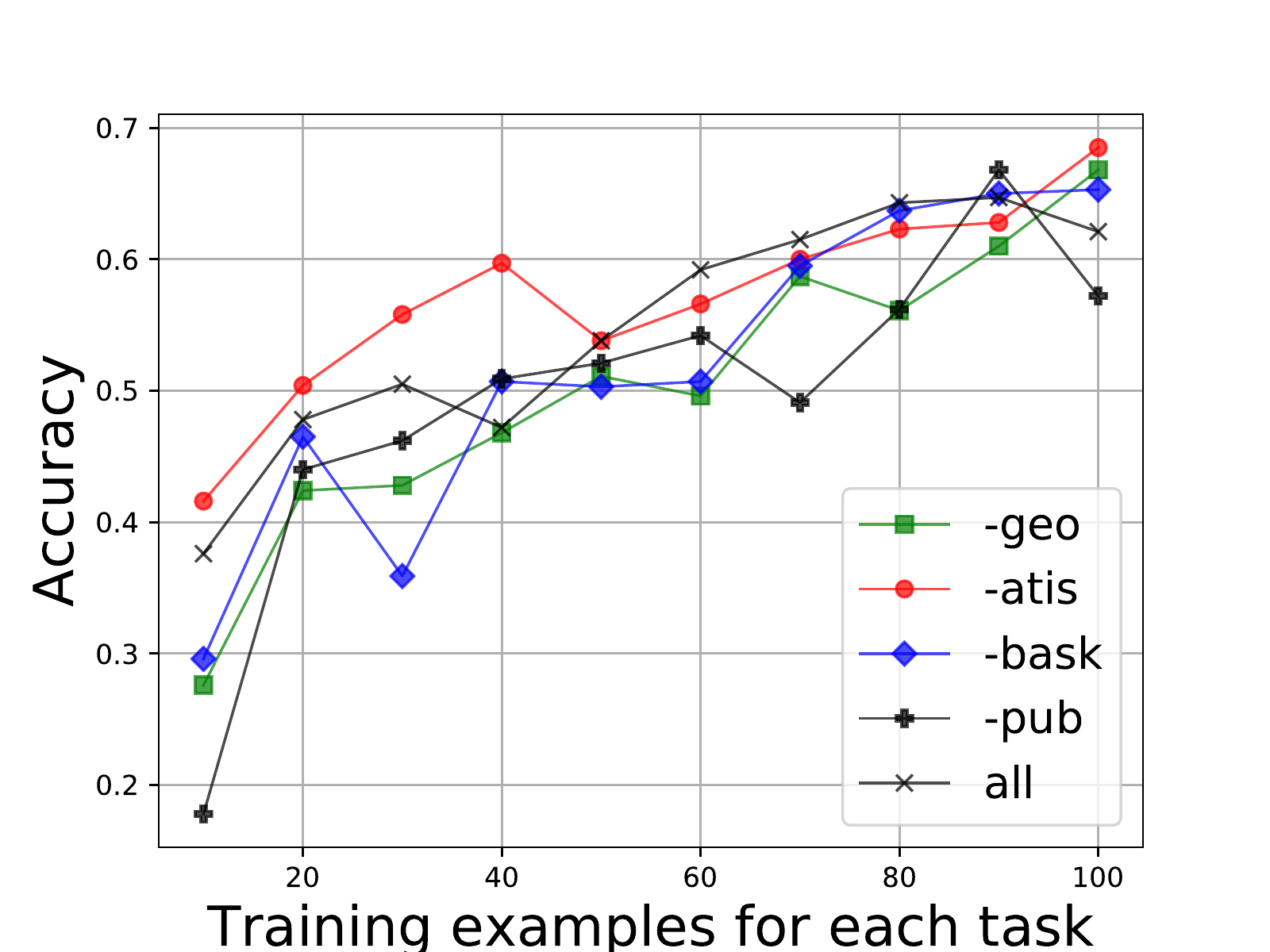}
        \caption{tok-level}
    \end{subfigure}
    \caption{Leave-one-out-domain results as a function of the number of training examples and the domains left out. As shown in the figures removing \texttt{pub} with few examples degraded the performance. An inverse outcome occurs when we remove \texttt{atis} with comparatively large number of examples.}
    \label{fig:leave-one-out}
\end{figure*}

As discussed in Section \ref{Influence Functions} $\mathcal{I}_{\uparrow,\mathcal{L}}(\textbf{z},z_{\textrm{test}})$ measures the impact of batch $\textbf{z} \in t_k$ on test examples. 
In this Section we evaluate this by $\mathcal{I}_{.,\mathcal{L}} - \mathcal{I}_{-\textbf{z},\mathcal{L}}$ which is leave-one-out cross-validation by definition \cite{Wasserman:2006}.
Table \ref{tab:res_table:leave-one-out} shows the experimental results on four different domains from the OVERNIGHT data set in which each time we leave one set out for testing on \texttt{calendar}.
Here we fixed $n=100$ to investigate the impact of each domain on \texttt{calendar}.
Fig. \ref{fig:leave-one-out} shows the parameter sensitivity of the method to the number of training points.

As shown in the figure, removing different $\textbf{z}$ causes varying effects on the performance of the system. 
For example, in the very initial steps of adding data for learning, the test set (\texttt{calendar}) takes more advantage of \texttt{publications} than others and removing this set of examples increases $\mathcal{L}$ considerably. 
This fact is highlighted more so in token-level accuracy, and this indicates that other examples propose useful tokens for prediction even though they might be in irrelevant classes/domains. 
However, \texttt{atis} is the only domain that does not provide useful information for token-level accuracy (e.g., $n<50$), and this is the noisiest domain in terms of denotation-level when the more number of examples are provided for train  (see the red hill in range $80-100$ in Fig. \ref{fig:leave-one-out} (a) and range $0-50$ in Fig. \ref{fig:leave-one-out} (b). It is clear from Fig. \ref{fig:leave-one-out} (b) that even this noisy domain provides useful tokens in range $80-100$.). 

\begin{table}[t]
\centering
\begin{tabular}{|c|c|c|c|c|}\cline{1-5}
test& out & seq & tok & den \\\hline
\multirow{5}{*}{ calendar } & all &  0.161& 0.621&0.262 \\ \cdashline{2-5}
&-geo & 0.149&0.668&0.220  \\ \cline{2-5}
& -atis & 0.143 & 0.685 & 0.196\\\cline{2-5}
& -bask & 0.083 & 0.653 & 0.238  \\ \cline{2-5}
& -pub & \textbf{0.077} & \textbf{0.572}  &\textbf{0.167}  \\ \cline{1-5}
\end{tabular}
\caption{Sequence-level, token-level and denotation-level accuracies of leave-one-out domain experiments. The column out refers to the domain left out and all means training and testing on only calendar data. The bold numbers belong to the cases where we see the highest drop in accuracy.}
\label{tab:res_table:leave-one-out}
\end{table}
\subsubsection{Label flipping}
\begin{figure}[th]
    \centering
    ~
    \begin{subfigure}[t]{0.22\textwidth}
        \centering
        \includegraphics[width=\linewidth]{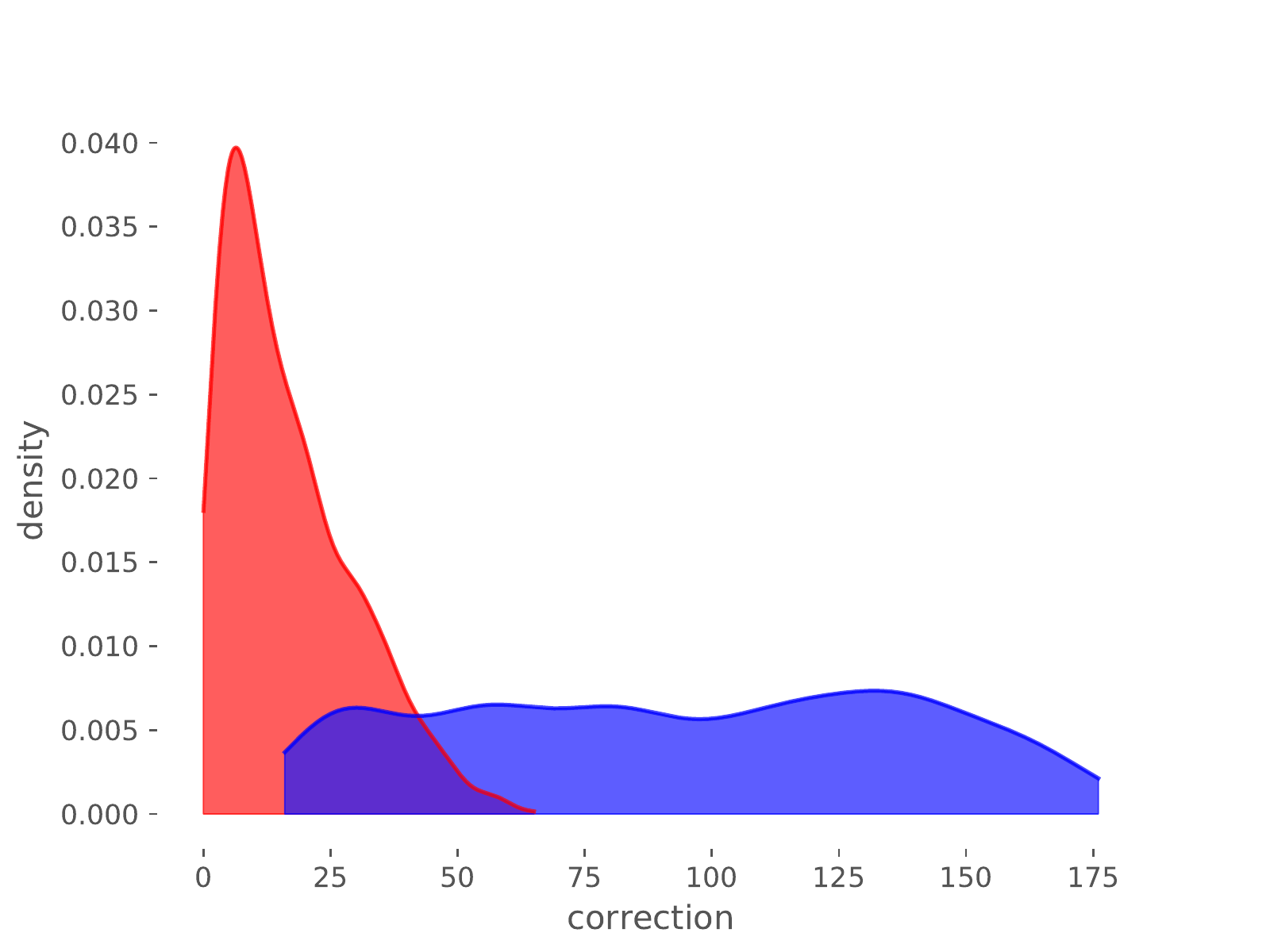}
        \caption{\texttt{cal+atis}}
    \end{subfigure}
        ~
    \begin{subfigure}[t]{0.22\textwidth}
        \centering
        \includegraphics[width=\linewidth]{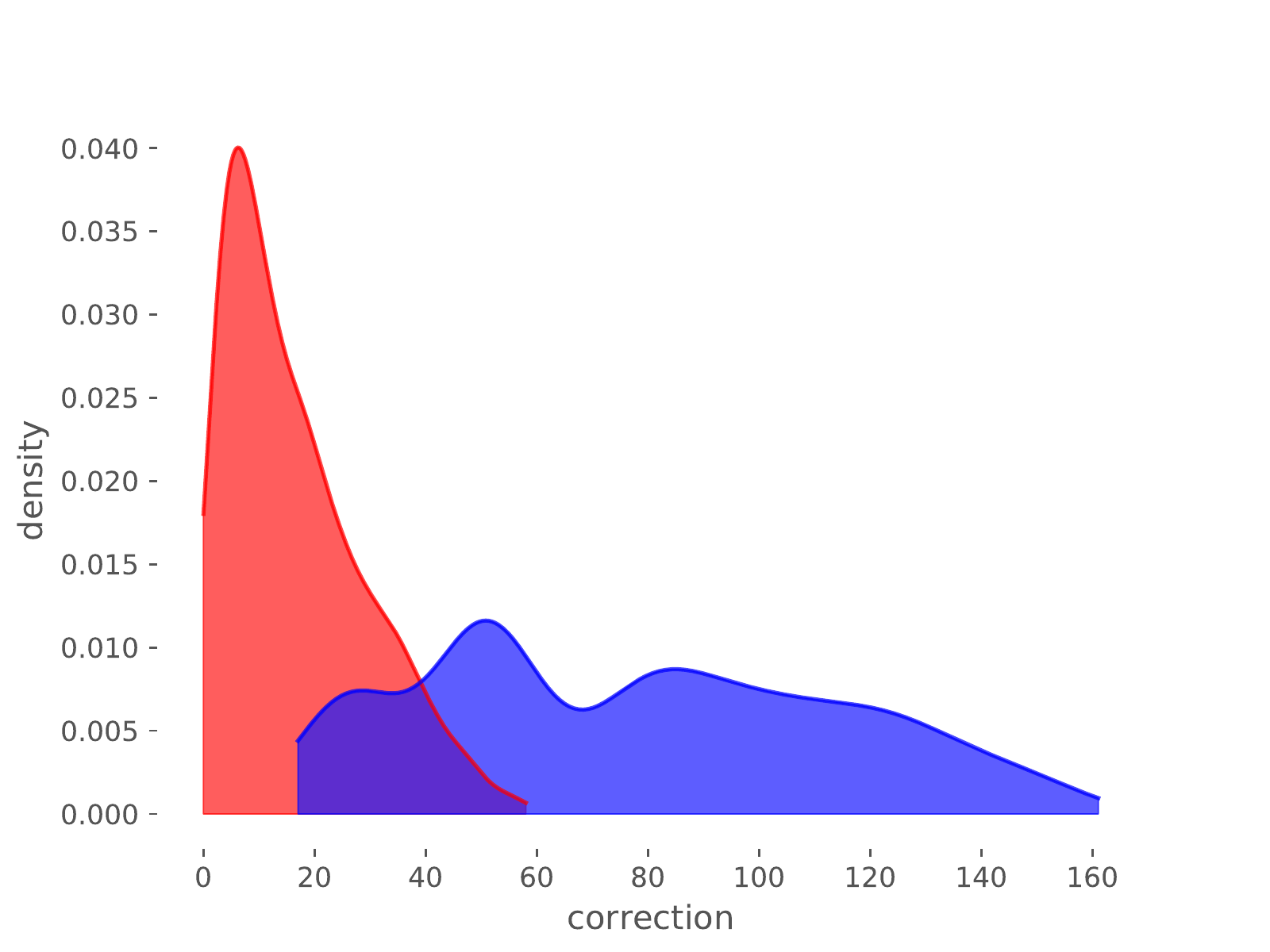}
        \caption{\texttt{cal+pub}}
    \end{subfigure}
    \caption{ Red curves indicate the average number of times random  selection detects mislabeled adversarial samples and the blue curves indicate average number of times influence functions found these examples. The horizontal axis shows the successfully trials and the vertical axis show the expected value.
    }
    \label{fig:random-hvp}
\end{figure}

In this section we conduct a set of experiments to visualize and find the most influential examples for domain classification.
A similar experiment is done by \cite{Koh:2017} for spam detection and image classification.
Here we set up multiple binary regression based on the Broyden\textemdash Fletcher\textemdash Goldfarb\textemdash Shanno algorithm for HVP estimation discussed in Section \ref{Influence Functions}.
The main goal is to find an efficient iterative solution for estimating the hessian-vector product in order to calculate the influence function.
To perform an adversarial attack we flipped the domain label of examples $z_j$ extracted from $\{0.05,0.1,..,0.25\}$ of the size of training data.
Later we want to study the correlation of the binary classification error in semantic parsing.
Fig. \ref{fig:random-hvp} shows the effectiveness of HVP in finding the most sensitive examples. 

As an additional test, we flipped the labels of \texttt{atis} samples to \texttt{calendar} and then measured the effectiveness of the influence function in detecting these label flips. We found that the distributions of the changes in $\mathcal{I}_{\uparrow,\mathcal{L}}(z_j,z_{\textrm{test}})$ or the classification loss and also the number of corrections are almost the same for both \texttt{publications} and \texttt{calendar} attacks. On the other hand, adverserial attacks on \texttt{atis} and \texttt{calendar} show different results; attacks on \texttt{calendar} result in a much lower number of flips. We interpret this result as evidence  that textual cross-attacking is not a bidirectional process, and sometimes adversarial attacks from one domain (e.g., \texttt{atis}) are detected and corrected more more easily (i.e., ranking the examples by their changes in the loss function following by HVPs and correcting the top$-k$ ones).



\subsubsection{Data augmentation}
\begin{figure*}[t]
    \centering
    ~
        \begin{subfigure}[t]{0.44\textwidth}
        \centering
        \includegraphics[width=\linewidth]{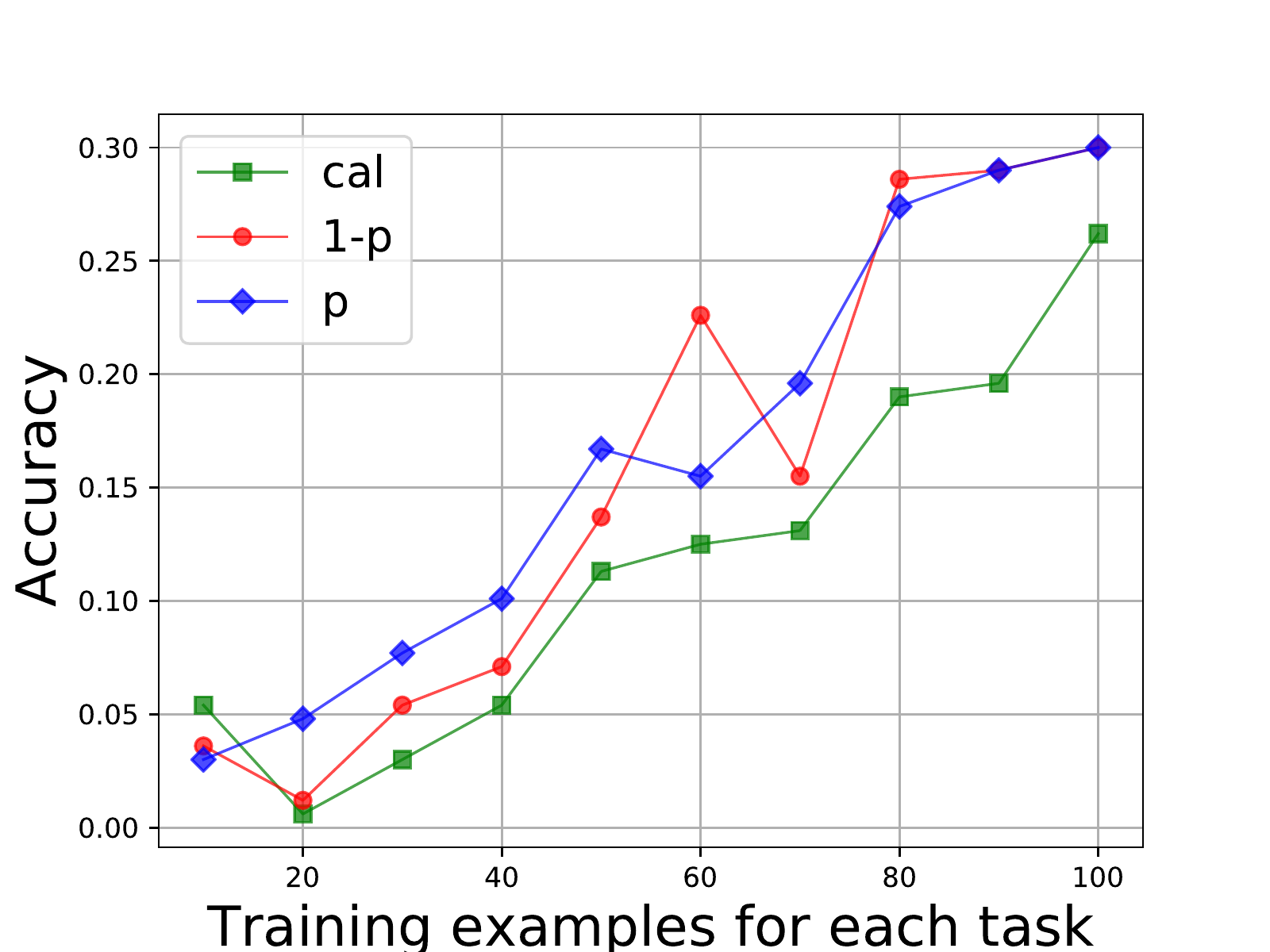}
        \caption{\texttt{pub}:den}
        \label{fig:cross-attack:a}
    \end{subfigure}
    ~    
    \begin{subfigure}[t]{0.44\textwidth}
        \centering
        \includegraphics[width=\linewidth]{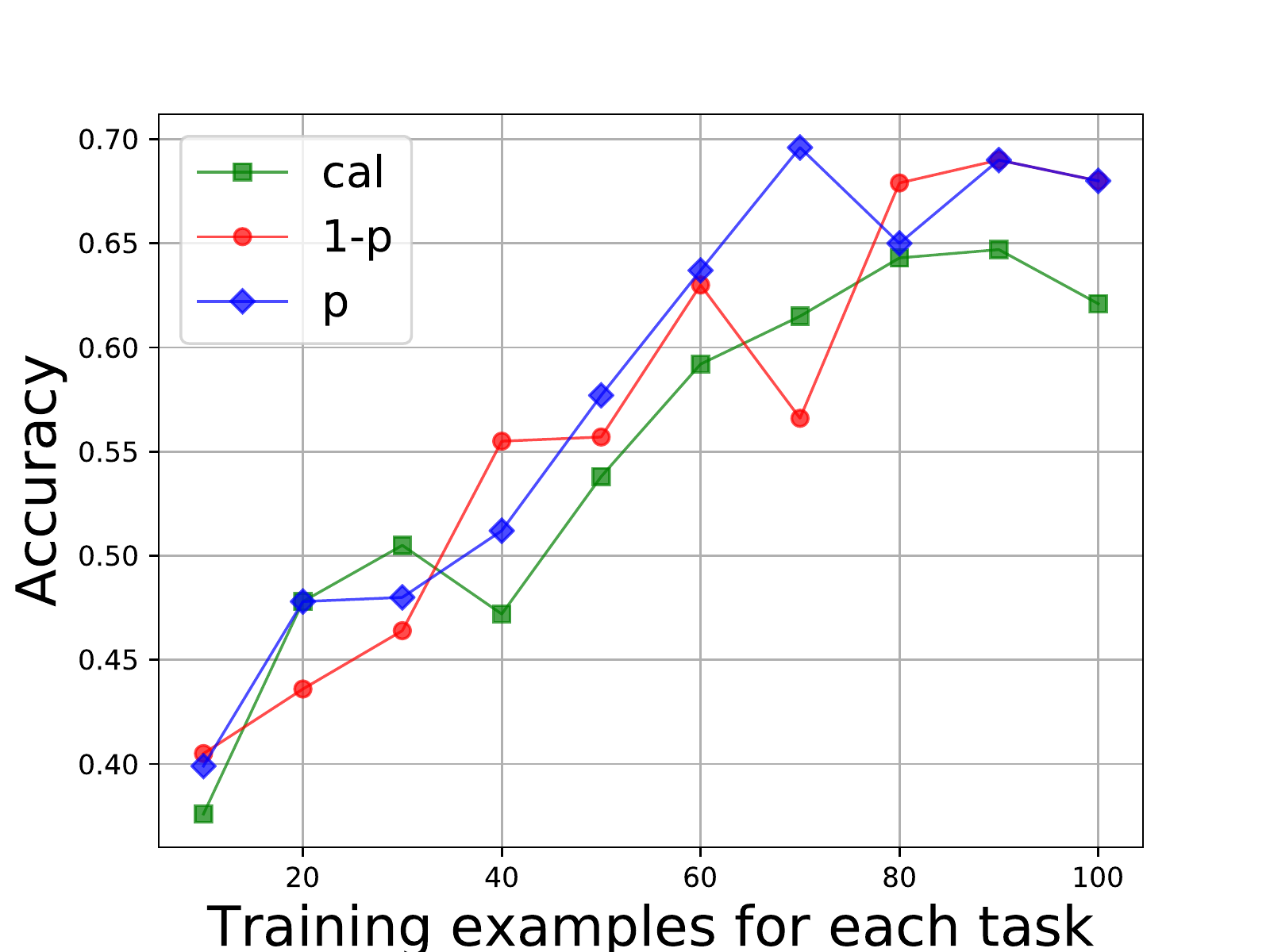}
        \caption{\texttt{pub}:tok}
         \label{fig:cross-attack:b}
    \end{subfigure}
    ~
        \begin{subfigure}[t]{0.44\textwidth}
        \centering
        \includegraphics[width=\linewidth]{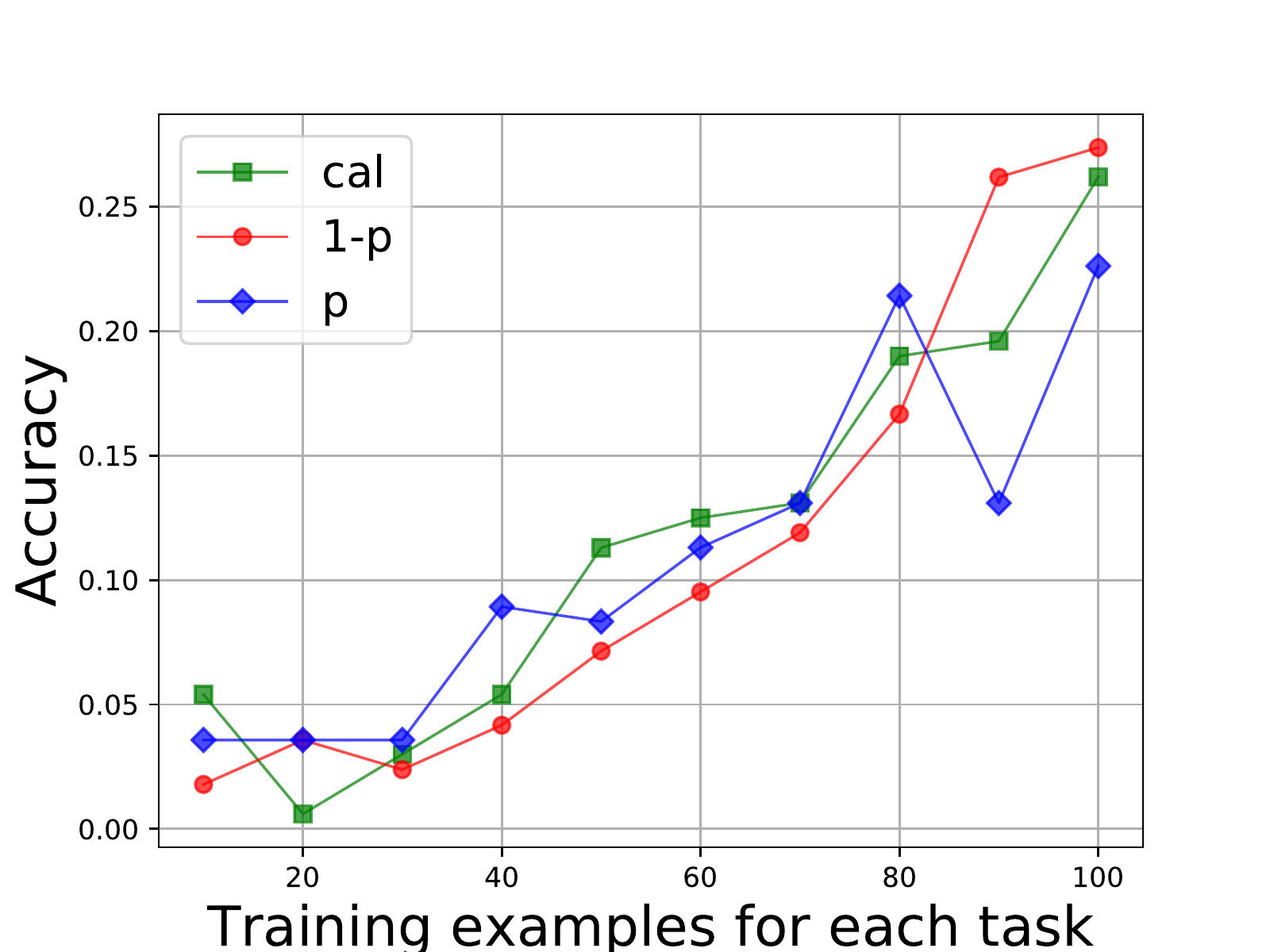}
        \caption{\texttt{atis}:den}
         \label{fig:cross-attack:c}
    \end{subfigure}
    ~    
    \begin{subfigure}[t]{0.44\textwidth}
        \centering
        \includegraphics[width=\linewidth]{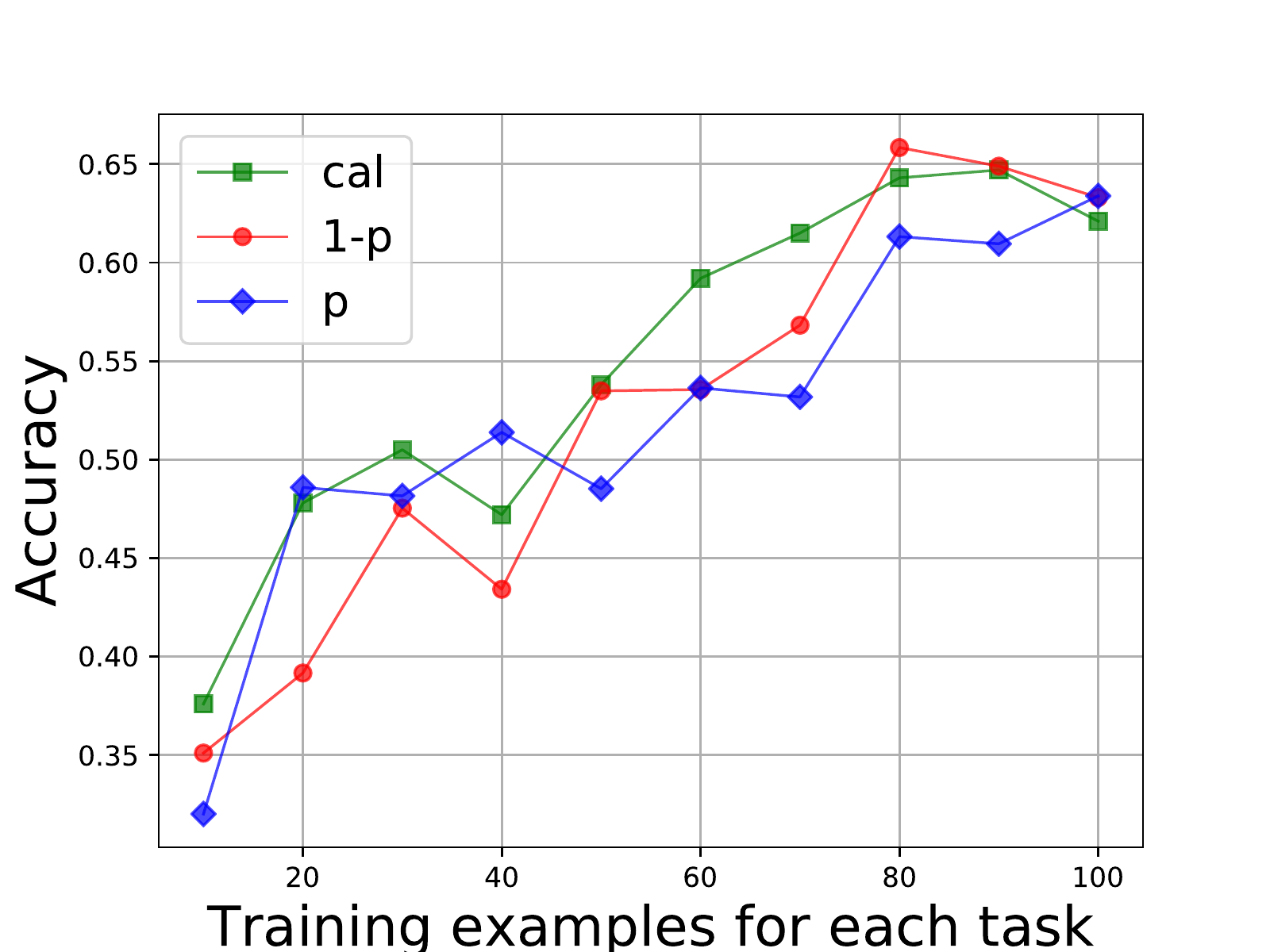}
        \caption{\texttt{atis}:tok}
         \label{fig:cross-attack:d}
    \end{subfigure}

    \caption{Adversarial sampling on \texttt{atis}, the least relevant domain for \texttt{calendar} (green) and \texttt{publications}, the most relevant one, followed by augmenting training with sampling according to $p$ (blue) and $1-p$ (red).}
    \label{fig:cross-attack}
\end{figure*}
In this part we attempt to improve semantic parsing performance by augmenting our training data with examples from other $t_{k'\neq k}$ to $t_k$ by extracting the most sensitive examples according to the influence function after running the HVP algorithm. 
To this aim we applied several adversarial attacks on $\texttt{atis}$ and $\texttt{publications}$ and checked the top $\{0.05,0.1,..,0.25\}$ samples of training data to select the most frequent examples. 
To this aim, we created distributions $p$ and $1-p$ based on how many times an example appeared in top list of HVPs for correction.
Our intuition is that samples being identified as influential examples for binary domain classification might be useful for the original sequential classification problem. 

We then added examples after $100$ sampling on $p$ and $1-p$  as training for $t_k=$\texttt{calendar} and tested on this domain.
As shown in Fig. \ref{fig:cross-attack} cross-attacking between \texttt{atis} and \texttt{calendar} is also useful in semantic parsing for the test examples in \texttt{calendar} when we have $n\leq 40$ examples (see blue line of $p$ in Fig. \ref{fig:cross-attack:c}).
On the other hand, examples from $1-p$ identifies rare examples which are also sensitive, and it is clear from the red line that examples from this distribution works well when we add more data points or increase the likelihood of noise. 
Figures~\ref{fig:cross-attack:a} and~\ref{fig:cross-attack:b} also support the considerable correlation between \texttt{publications} and \texttt{calendar} for which we see the similar outcome in Fig. \ref{fig:leave-one-out} in the leave-one-out experiment. 

\subsubsection{Number of parameters}
Table \ref{tab:parameters} shows the number of parameters introduced by multiple transfer learning approaches.
As shown in the table, o2o/e2d/zero-shot k*d use the fewest number of parameters. The o2o baseline provides poor performance in our experiments (see Fig.,\ref{fig:sing-agg}).
The model e2d uses a one-hot-augmented vector to control the parameters while our method, zero-shot k*d, defines more restrictive controls by slicing the $k-$th row of $W_T$ and providing a regularization constraint\footnote{e2d does not learn the augmented vector and zero-shot k*d slices the $k-$th row for each example. Therefore both learn $dV$ parameters for decoding. }.

\begin{table}[h]
\centering
\scalebox{0.72}{
\begin{tabular}{lll}
\cline{1-3}
method & \#parameters & reference\\\cline{1-3}
o2o &$\mathcal{O}(d^2+dV)$& \citet{Johnson:2016}\\
o2m &$\mathcal{O}(d^2+kdV)$& \citet{Fan:17}\\
m2m&$\mathcal{O}((k+1)d^2+kdV)$& \citet{Daume:2009}\\
e2d&$\mathcal{O}(d^2+dV)$&\citet{Herzig:17} \\
z-shot k*d &$\mathcal{O}(d^2+dV)$& Our Method\\
\cline{1-3}
\end{tabular}}
\caption{A comparison of the number of training parameters for multiple transfer learning models for zero-shot learning, where $d$ represents the size of the hidden dimension, $k$ the number of tasks and $V$ the size of the vocabulary.}
\label{tab:parameters}
\end{table}

\section{Conclusion and Future Work}
In this paper we experimented with zero-shot learning for semantic parsing.
Zero-shot learning in this context tries to maximize the the probability of predicting the current domain as a gold domain through a mapping matrix and minimizing the difference via a regularization term during decoding.
This formulation provides flexibility for sharing parameters between domains and controls the model from overfitting by introducing a fewer number of parameters on small datasets. 

We also conducted cross-domain adversarial attacks to augment our test data with the most influential examples.
Our experiments show that even the farthest domains contain useful information to be learned and further boosted the performance of our proposed model.


The results also support the view that in semantic parsing data sets cross-domain adversarial attack do not have a uniform effect, and influential functions detect more examples in one domain than the other one. In the future we plan to use this understanding to apply these models to more diverse datasets and tasks. 

\clearpage
\bibliography{ref}
\bibliographystyle{acl_natbib_nourl}

\clearpage
\appendix 
\section{Shared Encoder/Decoder}
\label{Shared Encoder/Decoder}
Let $x_t$ is the embedding at time step $t$ and $h_t$ as the hidden state, then:
\begin{eqnarray}
u_t = \textrm{tanh}(W_{xu}x_t+W_{hu}h_{t-1}+b_u) \nonumber\\
i_t = \sigma (W_{xi}x_t+W_{hi}h_{t-1}+b_i) \nonumber\\
o_t = \sigma(W_{xo}x_t+W_{ho}h_{t-1}+b_o) \nonumber\\
c_t = i_t \odot u_t + c_{t-1} \nonumber\\ 
h_t = o_t \odot \textrm{tanh}(c_t)
\end{eqnarray}
where $W_{xu}$ and $W_{hu}$ are the shared encoders between all training examples. 
$b_u$, $b_i$, and $b_o$ are bias, $i_t$ is the input gate, $o_t$ is the output gate, and $c_t$ is the context vector at time step $t$. 
The other option is to have a shared decoder between all the examples and define separate encoders for each domain:
\begin{eqnarray*}
  s_1=\textrm{tanh}(W^{(s)}[\overrightarrow{h}_m,\overrightarrow{h}^{k+1}_m,\overleftarrow{h}_1,\overleftarrow{h}^{k+1}_1])
\end{eqnarray*}
where $W^{(s)}$ is the decoder shared between all examples and there are $W^k_{x.}, W^k_{h.} ~\forall k, t_k \in T$ encoders for each task. 
Superscripts show forward and backward embedding respectively.

\section{Hessian vector product}
\label{hvp}
Suppose that $f(x)$ is a function, $g(x)=\frac{\partial f}{\partial x}$, $H(x)=\frac{\partial^2 f}{\partial^2 x}$ and  then we have:
\begin{eqnarray*}
      g(x+\Delta x) = g(x) + H(x)\Delta x +\mathcal{O}(x^2)\\
            g(x+\epsilon v) = g(x) + \epsilon H(x)v +\mathcal{O}(x^2) \\
            H(x)v = \frac{g(x+\epsilon v) - g(x)}{\epsilon} +\mathcal{O}(x^2) \\
            v= H^{-1}(x)\frac{g(x+\epsilon v) - g(x)}{\epsilon} \nonumber
\end{eqnarray*}
and also we have:
\begin{equation}
\begin{aligned}
      \mathcal{I}_{\uparrow,f}(x+\epsilon v)= &- \frac{\partial f}{ \partial \epsilon}\\
      =&-\frac{\partial f}{\partial x +\epsilon v}\frac{\partial x +\epsilon v}{\partial \epsilon} \nonumber\\
      =&-g(x+\epsilon v)^Tv \nonumber\\
      =&-g(x+\epsilon v)^TH^{-1}(x)\frac{g(x+\epsilon v) - g(x)}{\epsilon} \nonumber
\end{aligned}
\end{equation}
\end{document}